\documentclass[10pt,twocolumn,letterpaper]{article}

\usepackage[pagenumbers]{cvpr} 

\newcommand{\todo}[1]{{\color{red}#1}}
\newcommand{\TODO}[1]{\textbf{\color{red}[TODO: #1]}}

\definecolor{cvprblue}{rgb}{0.21,0.49,0.74}
\usepackage[pagebackref,breaklinks,colorlinks,allcolors=cvprblue]{hyperref}
\usepackage{colortbl} 
\usepackage{multirow} 
\usepackage{xcolor}   
\usepackage{array}    
\usepackage{graphicx}
\usepackage{caption}
\usepackage{algorithm}
\usepackage{algpseudocode}
\usepackage{fancyhdr}
\usepackage{amssymb}
\usepackage{placeins}
\def\paperID{14408} 
\def\confName{CVPR}
\def\confYear{2026}

\title{Personalized Image Descriptions from Attention Sequences}

\author{Ruoyu Xue\textsuperscript{1}, 
Hieu Le\textsuperscript{2},   Jingyi Xu\textsuperscript{1}, Sounak Mondal\textsuperscript{1}, 
Abe Leite\textsuperscript{1}\\
Gregory Zelinsky\textsuperscript{1}, Minh Hoai\textsuperscript{3}, Dimitris Samaras\textsuperscript{1} \\
\textsuperscript{1}Stony Brook University, USA~~~  
\textsuperscript{2}UNC-Charlotte, USA~~~ 
\textsuperscript{3}The University of Adelaide, Australia
}

\begin{document}
\maketitle

 \newcommand{\jx}[1]{{\color{red} #1}} 
\def\mA{\mathcal{A}}
\def\mB{\mathcal{B}}
\def\mC{\mathcal{C}}
\def\mD{\mathcal{D}}
\def\mE{\mathcal{E}}
\def\mF{\mathcal{F}}
\def\mG{\mathcal{G}}
\def\mH{\mathcal{H}}
\def\mI{\mathcal{I}}
\def\mJ{\mathcal{J}}
\def\mK{\mathcal{K}}
\def\mL{\mathcal{L}}
\def\mM{\mathcal{M}}
\def\mN{\mathcal{N}}
\def\mO{\mathcal{O}}
\def\mP{\mathcal{P}}
\def\mQ{\mathcal{Q}}
\def\mR{\mathcal{R}}
\def\mS{\mathcal{S}}
\def\mT{\mathcal{T}}
\def\mU{\mathcal{U}}
\def\mV{\mathcal{V}}
\def\mW{\mathcal{W}}
\def\mX{\mathcal{X}}
\def\mY{\mathcal{Y}}
\def\mZ{\mathcal{Z}} 

\def\bbN{\mathbb{N}} 
\def\bbR{\mathbb{R}} 
\def\bbP{\mathbb{P}} 
\def\bbQ{\mathbb{Q}} 
\def\bbE{\mathbb{E}}

\def\1n{\mathbf{1}_n}
\def\0{\mathbf{0}}
\def\1{\mathbf{1}}

\def\A{{\bf A}}
\def\B{{\bf B}}
\def\C{{\bf C}}
\def\D{{\bf D}}
\def\E{{\bf E}}
\def\F{{\bf F}}
\def\G{{\bf G}}
\def\H{{\bf H}}
\def\I{{\bf I}}
\def\J{{\bf J}}
\def\K{{\bf K}}
\def\L{{\bf L}}
\def\M{{\bf M}}
\def\N{{\bf N}}
\def\O{{\bf O}}
\def\P{{\bf P}}
\def\Q{{\bf Q}}
\def\R{{\bf R}}
\def\S{{\bf S}}
\def\T{{\bf T}}
\def\U{{\bf U}}
\def\V{{\bf V}}
\def\W{{\bf W}}
\def\X{{\bf X}}
\def\Y{{\bf Y}}
\def\Z{{\bf Z}}

\def\a{{\bf a}}
\def\b{{\bf b}}
\def\c{{\bf c}}
\def\d{{\bf d}}
\def\e{{\bf e}}
\def\f{{\bf f}}
\def\g{{\bf g}}
\def\h{{\bf h}}
\def\i{{\bf i}}
\def\j{{\bf j}}
\def\k{{\bf k}}
\def\l{{\bf l}}
\def\m{{\bf m}}
\def\n{{\bf n}}
\def\o{{\bf o}}
\def\p{{\bf p}}
\def\q{{\bf q}}
\def\r{{\bf r}}
\def\s{{\bf s}}
\def\t{{\bf t}}
\def\u{{\bf u}}
\def\v{{\bf v}}
\def\w{{\bf w}}
\def\x{{\bf x}}
\def\y{{\bf y}}
\def\z{{\bf z}}

\def\balpha{\mbox{\boldmath{$\alpha$}}}
\def\bbeta{\mbox{\boldmath{$\beta$}}}
\def\bdelta{\mbox{\boldmath{$\delta$}}}
\def\bgamma{\mbox{\boldmath{$\gamma$}}}
\def\blambda{\mbox{\boldmath{$\lambda$}}}
\def\bsigma{\mbox{\boldmath{$\sigma$}}}
\def\btheta{\mbox{\boldmath{$\theta$}}}
\def\bomega{\mbox{\boldmath{$\omega$}}}
\def\bxi{\mbox{\boldmath{$\xi$}}}
\def\bnu{\mbox{\boldmath{$\nu$}}}                                  
\def\bphi{\mbox{\boldmath{$\phi$}}}
\def\bmu{\mbox{\boldmath{$\mu$}}}

\def\bDelta{\mbox{\boldmath{$\Delta$}}}
\def\bOmega{\mbox{\boldmath{$\Omega$}}}
\def\bPhi{\mbox{\boldmath{$\Phi$}}}
\def\bLambda{\mbox{\boldmath{$\Lambda$}}}
\def\bSigma{\mbox{\boldmath{$\Sigma$}}}
\def\bGamma{\mbox{\boldmath{$\Gamma$}}}
                                  
\newcommand{\myprob}[1]{\mathop{\mathbb{P}}_{#1}}

\newcommand{\myexp}[1]{\mathop{\mathbb{E}}_{#1}}

\newcommand{\mydelta}[1]{1_{#1}}

\newcommand{\myminimum}[1]{\mathop{\textrm{minimum}}_{#1}}
\newcommand{\mymaximum}[1]{\mathop{\textrm{maximum}}_{#1}}    
\newcommand{\mymin}[1]{\mathop{\textrm{minimize}}_{#1}}
\newcommand{\mymax}[1]{\mathop{\textrm{maximize}}_{#1}}
\newcommand{\mymins}[1]{\mathop{\textrm{min.}}_{#1}}
\newcommand{\mymaxs}[1]{\mathop{\textrm{max.}}_{#1}}  
\newcommand{\myargmin}[1]{\mathop{\textrm{argmin}}_{#1}} 
\newcommand{\myargmax}[1]{\mathop{\textrm{argmax}}_{#1}} 
\newcommand{\myst}{\textrm{s.t. }}

\newcommand{\denselist}{\itemsep -1pt}
\newcommand{\sparselist}{\itemsep 1pt}

\definecolor{pink}{rgb}{0.9,0.5,0.5}
\definecolor{purple}{rgb}{0.5, 0.4, 0.8}   
\definecolor{gray}{rgb}{0.3, 0.3, 0.3}
\definecolor{mygreen}{rgb}{0.2, 0.6, 0.2}

\newcommand{\cyan}[1]{\textcolor{cyan}{#1}}
\newcommand{\blue}[1]{\textcolor{blue}{#1}}
\newcommand{\magenta}[1]{\textcolor{magenta}{#1}}
\newcommand{\pink}[1]{\textcolor{pink}{#1}}
\newcommand{\green}[1]{\textcolor{green}{#1}} 
\newcommand{\gray}[1]{\textcolor{gray}{#1}}    
\newcommand{\mygreen}[1]{\textcolor{mygreen}{#1}}    
\newcommand{\purple}[1]{\textcolor{purple}{#1}}       

\definecolor{greena}{rgb}{0.4, 0.5, 0.1}
\newcommand{\greena}[1]{\textcolor{greena}{#1}}

\definecolor{bluea}{rgb}{0, 0.4, 0.6}
\newcommand{\bluea}[1]{\textcolor{bluea}{#1}}
\definecolor{reda}{rgb}{0.6, 0.2, 0.1}
\newcommand{\reda}[1]{\textcolor{reda}{#1}}

\def\changemargin#1#2{\list{}{\rightmargin#2\leftmargin#1}\item[]}
\let\endchangemargin=\endlist
                                               
\newcommand{\cm}[1]{}

\newcommand{\mhoai}[1]{{\color{blue}\textbf{[MH: #1]}}}

\newcommand{\mtodo}[1]{{\color{red}$\blacksquare$\textbf{[TODO: #1]}}}
\newcommand{\myheading}[1]{\vspace{0.5ex}\noindent \textbf{#1}}
\newcommand{\htimesw}[2]{\mbox{$#1$$\times$$#2$}}

\newif\ifdraft
\drafttrue 
\definecolor{darkpink}{rgb}{0.91, 0.33, 0.5}
\definecolor{darkgreen}{rgb}{0.0, 0.5, 0.0}
\ifdraft
  \newcommand{\HL}[1]{{\color{orange}{\bf HL: #1}}} 
 \newcommand{\hl}[1]{{\color{orange} #1}}
 \newcommand{\RX}[1]{{\color{darkgreen}\textbf{[RX: #1]}}}
\else
  \newcommand{\HL}[1]{}
 \newcommand{\hl}[1]{#1}
 \newcommand{\ME}[1]{}
  \newcommand{\me}[1]{#1}
  \newcommand{\TODO}[1]{}
  \newcommand{\todo}[1]{#1}
\fi

\newcommand{\parag}[1]{\vspace{-3mm}\paragraph{#1}}


%
%
%

\newcommand{\Sref}[1]{Sec.~\ref{#1}}
\newcommand{\Eref}[1]{Eq.~(\ref{#1})}
\newcommand{\Fref}[1]{Fig.~\ref{#1}}
\newcommand{\Tref}[1]{Table~\ref{#1}}

\newcolumntype{C}[1]{>{\centering\arraybackslash}p{#1}}
\begin{abstract}
People can view the same image differently: they focus on different regions, objects, and details in varying orders and describe them in distinct linguistic styles. This leads to substantial variability in image descriptions. However, existing models for personalized image description generation focus on linguistic style alone, with no prior work leveraging individual viewing patterns. We address this gap by explicitly modeling personalized viewing behavior as a core factor in description generation. Our method, DEPER (DEscription–PERception persona encoder), learns a subject embedding that captures both linguistic style and viewing behavior, guided by an auxiliary attention-prediction task. A lightweight adapter aligns these embeddings with a frozen vision–language model, enabling few-shot personalization without retraining.
Across four datasets spanning diverse viewing tasks and both short and detailed descriptions, DEPER achieves a 24\% average improvement, showing that modeling personalized attention produces more human-aligned and high-quality descriptions. We posit that understanding how people see helps predict what they say; modeling human diversity in perception can improve both performance and human alignment in multi-modal systems.
\end{abstract}    
\section{Introduction}
\label{sec:intro}

\begin{figure}[t!]
  \centering
    \makebox[0.48\linewidth]{Subject A}
  \makebox[0.48\linewidth]{Subject B}
  \includegraphics[trim={0 0 0.3cm 0},width=1.0\linewidth]{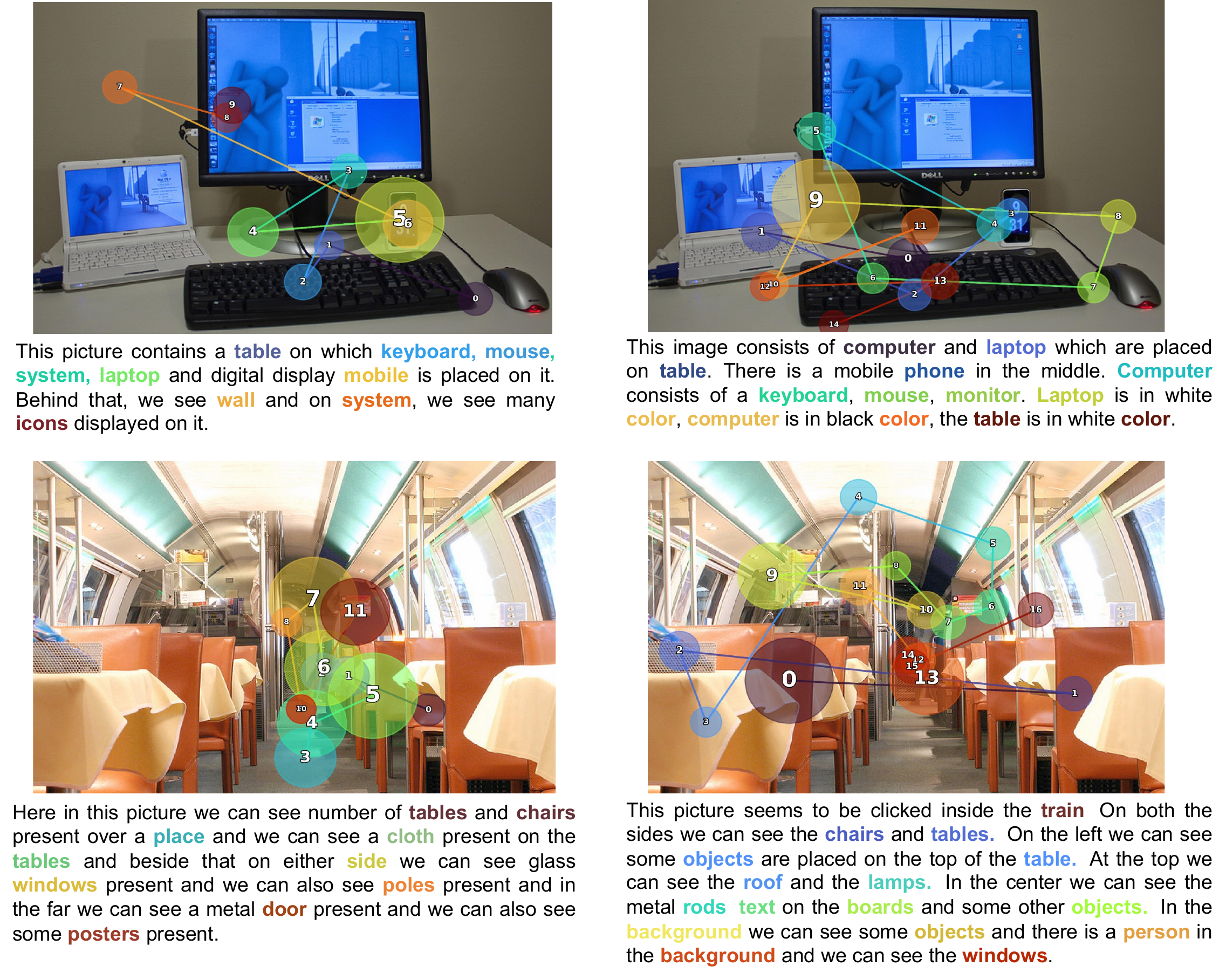}

  \caption{People have distinct viewing habits~\cite{chen2024beyond,xue2025few} that shape how they describe an image. Subject A moves between major objects, while Subject B inspects them in detail and in a different order. Our method models these patterns to produce personalized image descriptions. The images and annotations are from \cite{pont2020connecting}. }
  \vspace{-1.2em}
  \label{fig:teaser}
\end{figure}

People can view the same image differently: they focus on different regions, objects, and details in different orders, and describe them in their own distinct linguistic styles. These individual differences naturally lead to variability in image descriptions. The personalized image description task aims to capture these differences by conditioning generation on \textit{who} is describing the image, rather than only on \textit{what} is depicted \cite{chunseong2017attend}. Such personalization has broad applications, including accessibility for people with low vision \cite{stangl2021going,jiang2024s}, media summarization \cite{shetty2024detailed}, education \cite{leotta2023evaluating}, and preference-based product descriptions in advertisements \cite{chen2019towards}. Beyond practical benefits, it  enables more diverse, human-aligned descriptions that enrich multimodal representations \cite{lavoie2024modeling}.

Existing personalization models focus mainly on different linguistic style attributes such as vocabulary or tone
\cite{chunseong2017attend,park2018towards,long2020cross,xiong2020towards,wang2023user}. However, beyond personal linguistic style, each person also exhibits a consistent and distinctive pattern in how they view images~\cite{xue2025few}. Cognitive studies have shown that visual attention patterns shape what people describe \cite{griffin2000eyes,gleitman2007give,coco2012scan}, and numerous deep-learning models have further demonstrated their critical role in image description generation task \cite{alahmadi2022improve,takmaz2020generating,chen2018boosted,cornia2018paying,zhou2019re}. (see Fig. \ref{fig:teaser}). Yet, no prior work has incorporated attention patterns into personalization.

To this end, we propose DEPER (\underline{DE}scription-\underline{PER}cep\-tion persona encoder), a framework that grounds personalization in human attention. DEPER explicitly learns how individuals perceive and explore visual scenes: where they look, in what order, and for how long, and uses these attention patterns to guide personalized image descriptions. There are two key challenges in leveraging personalized attention into description generation: (1) Modeling personalized human attention is difficult because human attention is noisy, continuous, and behaviorally diverse. Moreover, since attention is strongly conditioned on image content, disentangling stable personalized attention traits from content-specific cues is non-trivial \cite{xue2025few,chen2024beyond}. (2) Adapting image description models to new subjects with few support examples easily leads to overfitting because they are usually parameter-heavy. In real-world deployment, the model must rapidly adapt to and capture user-specific preferences from limited interaction data for new subjects.

The central component of DEPER is a content-invariant subject embedding that captures a viewer’s characteristic patterns of visual attention and linguistic expression.
Specifically, DEPER employs a dual-context encoder that jointly models visual trajectories and the corresponding image descriptions. A subject embedding extractor distills this multi-modal information into the subject embedding, while an auxiliary attention trajectory reconstruction objective encourages the embedding to retain key attention dynamics. To generate personalized outputs, an adapter maps the subject embedding into the embedding space of a pretrained vision-language model (VLM). This allows us to prompt the VLM with \textit{``Write a description of this photo in the style of \texttt{<subj>}.''}, where \texttt{<subj>} is the subject-specific token.

During inference, DEPER can produce personalized image descriptions without requiring attention data, since the model has already internalized individual attention patterns within the subject embedding. Additionally, a contrastive loss disentangles subject embeddings from image content, enabling efficient adaptation to new subjects from only a few examples, without additional fine-tuning, while remaining memory- and time-efficient for real-time applications.


We evaluate on four datasets \cite{pont2020connecting,he2019human,kollenda2025individual} that measure attention either via mouse movements or human gaze collected by an eye-tracker, covering three distinct viewing tasks and including either concise or detailed image descriptions.
Across these settings, DEPER consistently outperforms baselines, improving BLEU-4 \cite{papineni2002bleu} by 12\% and CIDEr \cite{vedantam2015cider} by 20\% on average for both concise and detailed image description tasks.

In summary, our contributions are:

\begin{itemize}
    \item We develop DEPER, the first method to personalize image description by leveraging human attention.
    \item Through extensive experiments, we show that DEPER reliably achieves superior performance versus baselines.
    \item We demonstrate that DEPER can adapt to new subjects with few supporting samples (few-shot personalization).
\end{itemize}

\section{Related Work}
\label{sec:related-work}
\vspace{3mm}
\parag{Personalized Image Captioning.}
Personalized image captioning was introduced by CSMN~\cite{chunseong2017attend}, which predicts captions from Instagram posts by building a user-context memory via TF–IDF~\cite{salton1988term} over a user's prior posts. This treats the most frequently used words of an individual in the past as the representation of their personality. Subsequent work \cite{park2018towards,long2020cross,xiong2020towards,wang2023user} largely retains this TF–IDF representation with architectural improvements. MHTN \cite{zhang2020learning} and UMCap \cite{nguyen2024umcap} add short- and long-term user representations or key-value maps to capture literal preference rather than TF-IDF, but their representation of personality still remains grounded in frequently used words within recurring contexts. A related task \cite{shuster2019engaging,hosseinzadeh2023few,zeng2019automatic,wang2022image,wang2018social,wang2023caption,gan2017stylenet} involves style-controlled caption generation via explicit text conditioning, such as ``sweet'' or ``dramatic''.

However, prior approaches, with or without explicit style control, generally {\it ignore human visual attention}, which is a key driver of what gets mentioned, in what order ,and the amount of details. Further, they mostly target short captions, while detailed, personalized descriptions remain underexplored. Our work is the first to treat personalized human attention as a key signal for image description generation, and we demonstrate consistent gains for both short and detailed description generation tasks.

\parag{Human Attention in Image Description.}
Human attention has been used to improve description models \cite{takmaz2020generating,chen2018boosted,cornia2018paying,zhou2019re}, typically as an auxiliary signal predicted by pretrained models\cite{chen2018boosted,cornia2018paying,zhou2019re}, sourced from ground-truth gaze\cite{takmaz2020generating}, or approximated by noun-aligned boxes \cite{alahmadi2022improve}. But these signals reflect population-level viewing tendencies, not individual traits. Attention-controlled methods \cite{wang2023caption,yan2021control,wang2024g,feng2022human,meng2021connecting} generate descriptions conditioned on fixation cues, yielding spatial–temporal interpretations aligned with the given gaze. Yet they do not learn subject-specific preferences that generalize across images and therefore require attention input at inference, which is impractical for large-scale use. Overall, prior work treats attention as an image-specific signal, not as a transferable, subject-level preference. In contrast, we learn personalization jointly from image, description, and attention signals, capturing individual viewing styles that persist across images (e.g., preference for background vs. foreground, or people vs. actions).

\parag{Identity-Based Personalization in Vision-Language Models.}

A common form of VLM personalization learns ``identity tokens'', embeddings that encode a subject identity. Subject-driven text-to-image models like DreamBooth \cite{ruiz2023dreambooth} map a person or object's appearance into such tokens to generate images containing this subject, and personalized retrieval systems like \cite{yeh2023meta} use such tokens to localize the subject across new contexts. Recent personalized VLM assistants \cite{pham2025plvm, nguyen2024yo, hao2025rap, nguyen2025yo, ryan2025improving, alaluf2024myvlm, pi2024personalized} similarly ground subject identity so the model can recognize and answer questions about a specific depicted entity. For instance, Yo'LLaVA \cite{nguyen2024yo} learns latent tokens representing a pet like ``bo'', enabling subject-specific dialogue. These methods equate personalization with identity recognition, emphasizing appearance and explicit attributes.
In contrast, we personalize to a viewer's latent viewing patterns as well as their linguistic tendencies. 
\section{Method}
\begin{figure*}[t]
  \centering
  \includegraphics[width=1.0\linewidth]{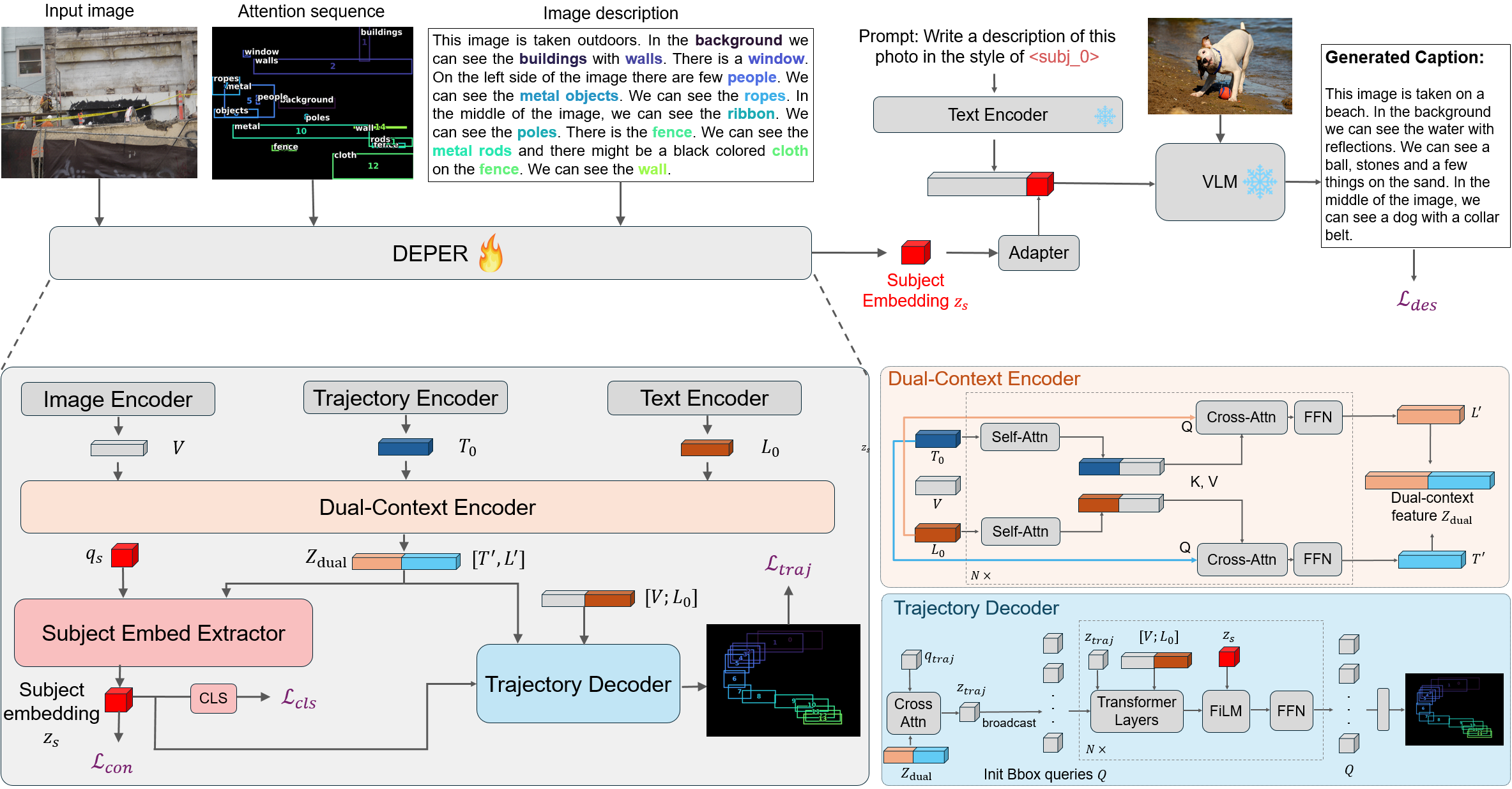}
  \caption{\textbf{Overview of DEPER and Caption Generation}: DEPER extracts a subject embedding $\mathbf{z}_s$ from a triplet ($I, D_s, T_s$), capturing the personalized viewing patterns and linguistic style. $\mathbf{z}_s$ then conditions a VLM to produce subject-aligned image descriptions. A Dual-Context Encoder aligns perceptual and linguistic information into $\mathbf{Z}_\mathrm{dual}$. A Subject Embedding Extractor then distills $\mathbf{Z}_\mathrm{dual}$ to $\mathbf{z}_s$, yielding personalized attention–linguistic traits. $\mathbf{z}_s$ is distinctive across subjects yet consistent across images, enforced by classification and contrastive losses. A trajectory decoder further encourages $\mathbf{Z}_\mathrm{dual}$ to capture viewing dynamics, and helps $\mathbf{z}_s$ capture a subject's exploration behavior.}
  \vspace{-1em}
  \label{fig:pipeline}
\end{figure*}
We propose DEPER, a novel method for personalized image description generation. DEPER learns to represent each person as a \textit{subject embedding}, a vector that summarizes how they tend to look at and describe images. This embedding is learned from triplets of $(I, D_s, T_s)$ (the image, the subject’s description, and their attention trajectory) to capture stable, subject-specific patterns of how individuals describe images.
This is achieved through our DEPER network (\cref{sec:encoder}). 
An adapter then projects the subject embedding into the frozen VLM space (\cref{sec:VLM}), enabling description generation in that person’s style without requiring gaze input at test time. To our knowledge, this is the first framework to embed individual attention dynamics into a transferable subject representation for vision–language generation. 
The overall architecture is illustrated in \cref{fig:pipeline}.

\subsection{\underline{DE}scription-\underline{PER}ception Persona Encoder}
\label{sec:encoder}

DEPER is designed to learn a subject embedding $\mathbf{z}_s$ that serves as a personalized representation capturing both how a person perceives and describes visual scenes.
We require $\mathbf{z}_s$ to be consistent across images for a single viewer, distinctive across viewers, and structured in a way that reflects real behavioral tendencies. To achieve this, DEPER integrates three complementary modules: A \textbf{dual-context encoder} (\cref{sec:dual-context-encoder}) fuses visual, linguistic, and attentional streams to capture how perception and expression interact. 
A \textbf{subject embedding extractor} (\cref{sec:latent}) distills this fused representation into the compact, subject-specific $\mathbf{z}_s$ under discriminative supervision. A \textbf{ trajectory decoder} (\cref{sec:traj}) reconstructs each subject’s viewing sequence conditioned on $\mathbf{z}_s$, reinforcing its capacity to capture personalized attention dynamics.

\subsubsection{Dual-Context Encoder}
\label{sec:dual-context-encoder}
The dual-context encoder computes a representation capturing both perception and expression via two interacting streams that repeatedly exchange information through cross-attention.

Given an image $I$, description $D_s$, and trajectory $T_s = \{(b_i, \tau_i)\}_{i=1}^{M}$ (where each $\mathbf{b}_i$ is a bounding box with duration $\tau_i$.
we extract image patch features $\mathbf{V}$, text token features $\mathbf{L_0}$, 
and trajectory features $\mathbf{T_0}$ (See supplementary for the details of obtaining trajectory features). To better capture visual dynamics, especially the scan order of scene exploration, and the duration of attention on each region, we apply sinusoidal positional encoding \cite{dosovitskiy2020image} to encode the box duration and position (the index of boxes), and add them to trajectory features, yielding multi-scale temporal features. The encoder alternates self- and cross-attention across the two streams: text tokens attend to image and trajectory context, and trajectory tokens attend to image and text context:
\vskip -0.2in
\begin{align}
\mathbf{T_{\ell+1}} &= \mathrm{FFN}_{\ell_T}\big(\mathrm{Cross}_{\ell_T}(\mathrm{Self}_{\ell_T}(\mathbf{T_{\ell}}), [\mathbf{V}; \mathbf{L_{\ell}}])\big), \\
\mathbf{L_{\ell+1}} &= \mathrm{FFN}_{\ell_L}\big(\mathrm{Cross}_{\ell_L}(\mathrm{Self}_{\ell_L}(\mathbf{L_\ell}), [\mathbf{V}; \mathbf{T_\ell}])\big).
\end{align}
Here, $\mathrm{Self}(\cdot)$, $\mathrm{Cross}(\cdot)$ and $\mathrm{FFN}(\cdot)$ denote self-attention layer following the design in \cite{vaswani2017attention}, $[;]$ is concatenation. We repeat it for $\ell$ layers. This design yields a fused representation $\mathbf{Z}_{\text{dual}} = [\mathbf{L}'; \mathbf{T}']$ that inherently captures each subject’s linguistic style and attentional behavior.

\subsubsection{Subject Embedding Extractor} 
\label{sec:latent}

We design the subject embedding extractor to produce a representation that captures a person’s consistent perceptual–linguistic behavior. 
It consists a learnable subject query $q_s$ that attends to $\mathbf{Z}_{\text{dual}}$ through a cross-attention layer. The subject query is optimized to selectively aggregate features that capture a specific subject's distinctive viewing and descriptive patterns across images, resulting in a stable subject embedding $\mathbf{z}_{s}$. 
To this end, we supervise the extractor with a joint objective that integrates subject classification and supervised contrastive learning \cite{khosla2020supervised}. This joint formulation strengthens representational discriminability and prevents collapse into a shared subspace.


\myheading{Subject embedding Losses. } A classification head predicts the subject ID from $\mathbf{z}_s$ to enforce inter-subject discrimination, optimized using a standard cross-entropy loss ($\mathcal{L}_{cls}$). An additional contrastive loss ($\mathcal{L}_{con}$, SupCon~\cite{khosla2020supervised}) pulls embeddings of the same subject closer and pushes those of different subjects apart. See details in the supplementary.

\subsubsection{Trajectory Decoder}
\label{sec:traj}
A key component of our framework is the integration of personal viewing patterns into the subject embedding. 
To enforce this, we introduce trajectory decoder to reconstruct the attention trajectory. This decoder takes as input an instance-specific trajectory latent $\mathbf{z}_{traj}$, and reconstructs that instance's attention trajectory $T_s$ while being conditioned on the visual features $\mathbf{V}$, description features $\mathbf{L_0}$, and subject embedding $\mathbf{z}_s$ in a personality-aware manner. 
Thus, $\mathbf{z}_s$ learns stable, individual viewing patterns that guide trajectory reconstruction, without collapsing into instance-specific memorization.

\myheading{Decoding block.} We initialize a learnable trajectory query $\mathbf{q}_{\mathrm{traj}}$ that extracts instance-specific attention dynamics from $\mathbf{Z}_{\text{dual}}$ through cross-attention, yielding the trajectory latent $\mathbf{z}_{\mathrm{traj}}$. 
We then initialize a sequence of $M$ box queries $Q_0=\{\mathbf{q}_i\}_{i=1}^{M}$ and broadcast $\mathbf{z}_{traj}$ to each query as a global prior. 
Each layer first applies self-attention within the box queries, followed by cross-attention to $\mathbf{z}_{traj}$, providing step-wise latent control throughout decoding:
\begin{align}
\mathbf{Q_{\ell}} &=
\mathrm{Cross}_\ell\!\Big(
    \mathrm{Self}_\ell\!\big(\mathbf{Q_\ell}\big),\,
    \mathbf{z}_{traj}
\Big).
\label{eq:decoder1}
\end{align}

We then condition the reconstruction on the visual and linguistic context via cross-attention to the image features $\mathbf{V}$ and description features $\mathbf{L_0}$. We use FiLM~\cite{perez2018film} with the subject embedding $\mathbf{z}_s$ so that it modulates each decoder block in a personality-aware manner. Finally, we apply a position-wise feed-forward network:
\begin{align}
\mathbf{Q_{\ell+1}} &=
\mathrm{FFN}_\ell\!\Big(\mathrm{FiLM}\!\Big(\mathrm{Cross}_{\ell}\!\Big(
    \mathbf{Q_\ell},\,
    [\mathbf{V};\mathbf{L_0}]
\Big), z_s\Big)\Big).
\label{eq:decoder2}
\end{align}

We repeat it for $\ell$ layers to get final box embeddings $Q'$. Last, we apply a linear head to the $\mathbf{Q'}$ to obtain the box coordinates $(\hat{x}_{\min},\hat{y}_{\min},\hat{x}_{\max},\hat{y}_{\max},\hat{v})$, yielding the predicted box and validity sequences $\hat{\mathbf{B}} \in [0,1]^{T\times 4}$ and $\hat{\mathbf{V}} \in [0,1]^T$.

\myheading{Training Objective.} We use a smooth L1 loss ($\mathrm{SL_1}$) on the box predictions $\hat{\mathbf{B}}$, and a binary cross-entropy loss (BCE) on $\hat{\mathbf{V}}$, for an overall loss $\mathcal{L}_{\text{traj}} = \mathcal{L}_{\text{box}} + \mathcal{L}_{\text{valid}}$:
\vspace{-0.8em}
\begin{align}
& \mathcal{L}_{\text{box}}
= \big(\sum_{i=1}^{M} v_i\, \mathrm{SL_1}(\hat{\mathbf{b}}_i,\mathbf{b}_i)\big)
  \big/ \big(\sum_{i=1}^{M} v_i + \varepsilon\big), \\
& \mathcal{L}_{\text{valid}}= \frac{1}{M}\sum_{t=1}^{M} \mathrm{BCE}(\hat v_i,v_i). \nonumber 
\end{align}

\subsection{Image Description Generation}
\label{sec:VLM}

\myheading{VLM conditioning.} Given a subject embedding, our next goal is to make the VLM generate descriptions in that person’s style. We achieve this by injecting the embedding directly into the model’s prompt space. First, we add subject tokens \texttt{<subj\_x>} into VLM's vocabulary (\texttt{x} denotes different subjects). Then a small adapter (single linear layer, as in LLaVA \cite{liu2023visual}) maps the DEPER embedding into the VLM’s token dimension, and then this adapted vector replaces the embedding of a dedicated subject token in the prompt. The rest of the VLM remains frozen; it simply treats this subject vector as part of the input sequence and adapts its generation accordingly. To be precise, in the prompt ``\textit{Write a description of this photo in the style of \texttt{<subj\_x>}.}'', we replace the token \texttt{<subj\_x>} with the adapted subject vector at the embedding layer. This gives the model a direct, continuous representation of the subject’s style while keeping the prompt and VLM architecture unchanged. To avoid information leakage since DEPER takes descriptions as input, we condition the VLM on a \emph{different} image–description pair \((I',D_s')\) from the same subject as the \((I,D_s,T_s)\) used by DEPER. 

\myheading{Captioning Loss. } We use Supervised Fine-Tuning \cite{ouyang2022training}: the VLM generates $\hat{D_s'}$ of length $N_{D'}$ conditioned on prompt $P(\mathbf{z}_s)$:
\vspace{-0.8em}
\begin{align}
\mathcal{L}_{\text{des}}
= - \sum_{t=1}^{N_{D'}} \log p_\phi\!\big(d_t \,\big|\, I,\; P(\mathbf{z}_s),\; d_{1\cdots t-1}\big)
\end{align} 



\subsection{Training}
\label{sec:train-objective}
We train DEPER in two stages. In Stage 1, we train with $\mathcal{L}_{\text{stage1}}=\lambda\mathcal{L}_{\text{con}}+\mathcal{L}_{\text{traj}}+\mathcal{L}_{\text{cls}}$. $\mathcal{L}_{\text{des}}$ and $\mathcal{L}_{\text{cls}}$ supervise DEPER to optimize subject embeddings to be image-independent and personality-aware; $\mathcal{L}_{\text{traj}}$ forces the dual-context encoder to learn visual dynamics.

In Stage 2, we freeze the dual-context encoder as it has been trained to a robust representation. We train the subject embedding extractor and the VLM adapter to utilize the pretrained knowledge in Stage 1, and align subject embeddings to VLM space. The training loss is $\mathcal{L}_{\text{Stage2}}=\mathcal{L}_{\text{des}}+\lambda\mathcal{L}_{\text{con}}+\mathcal{L}_{\text{cls}}$. The classification and contrastive losses in Stage 2 ensure that subject embeddings remain distinct and stable in the subject space.

\section{Experiments}
\label{sec:experiments}


\myheading{Datasets.} We evaluate on four datasets. (1) Localized Narratives \cite{pont2020connecting}: participants describe images while moving the mouse over mentioned regions, yielding paired text–trace annotations on public image sets such as COCO \cite{lin2014microsoft} and Flickr30k \cite{plummer2015flickr30k}. We denote them as COCO-LN and Flickr30k-LN, with 108 and 37 subjects respectively. (2) He et al. \cite{he2019human}: 5 subjects view 1000 images and simultaneously speak a one-sentence caption. (3) Kollenda et al. \cite{kollenda2025individual}: participants observe 100 natural scenes for 3 s, then briefly describe them post-viewing, with gaze recorded during observation. For each dataset, we split training subjects into seen and unseen splits. Each unseen subject has 5 samples as the few-shot support set. We repeat this sampling by 5 times and report the average scores for reliability. Detailed statistics are shown in \cref{tab:data-stats}. Results on both seen and unseen are reported on the official test set when available; otherwise, a validation subset is sampled from the training data and the original validation set is used for testing.

\myheading{Baselines:} MITR\cite{meng2021connecting}: A population-level (without personalization) image description model to generate image descriptions on Localized Narratives with train-time attention trajectory supervision. Qwen Zero-shot: Qwen2-VL-2B\cite{wang2024qwen2vlenhancingvisionlanguagemodels} off-the-shelf. It generates descriptions for each image, without personalization. CSMN \cite{chunseong2017attend}: to our knowledge, the only personalized image captioning model with publicly released code. Qwen+PT: Qwen with prompt tuning \cite{lester2021power}. We add user-specific words (\texttt{<subj\_1>}, \texttt{<subj\_2>}, ...) to Qwen2-VL-2B\cite{wang2024qwen2vlenhancingvisionlanguagemodels}'s vocabulary and update only the corresponding weights in the input embedding layer that maps these tokens to embeddings, keeping all other parameters frozen. MITR-FT: MITR\cite{meng2021connecting} fine-tuned separately for each subject using all of that subject’s data. Qwen few-shot: We prompt Qwen using the query image and the support set’s descriptions and attention trajectories, guiding it to generate a description consistent with the stylistic patterns demonstrated in the examples. See supplementary for the prompt we use.

\begin{table}[t]
\caption{\textbf{Dataset statistics.} The datasets are different sizes, but all have low human consistency, showing high variability among subjects. [\#train/test]: number of annotated images; [\#seen/unseen]: number of subjects in train and few-shot sets; [length] avg. description length; [HC] human consistency (m-BLEU-4).}
\centering
\small
\setlength{\tabcolsep}{2.8pt}
\renewcommand{\arraystretch}{1.05}

\label{tab:data-stats}
\resizebox{0.8\columnwidth}{!}{
\begin{tabular}{l|c@{\hspace{1.5pt}}c@{\hspace{5.5pt}}c@{\hspace{2.5pt}}c@{\hspace{1.5pt}}c@{\hspace{1.5pt}}}
\toprule
\textbf{Dataset} & \textbf{\#train/test} & \textbf{\#seen/unseen} & \textbf{length} & \textbf{HC} \\
\midrule
\textbf{COCO-LN}\cite{pont2020connecting}      & 134272 / 8573 & 89 / 19 & 41 & 0.037 \\
\midrule
\textbf{Flk30k-LN}\cite{pont2020connecting}    & 30546 / 1023 & 27 / 10 & 57 & 0.061 \\
\midrule
\textbf{Kollenda et al.}\cite{kollenda2025individual} & 1950 / 450 & 22 / 8 & 16 & 0.054 \\
\midrule
\textbf{He et al.}\cite{he2019human}          & 3998 / 1000 & 5 / 0 & 8 & 0.054 \\
\bottomrule
\end{tabular}
}

\vskip -0.1in
\end{table}

\subsection{Implementation details. } 
\label{sec:implementation}

\myheading{Training and configuration details. }The hidden dimension of DEPER is set to 384. The dual-context encoder, subject embedding extractor, and trajectory decoder use 2, 1, and 4 layers, respectively. We set $\lambda = 0.1$ for the contrastive loss. DEPER is trained for 40 epochs in Stage 1 and 15 epochs in Stage 2 with a learning rate of 0.0005. Training is conducted on two RTX A6000 GPUs with a batch size of 16 and gradient accumulation step of 4. We use Qwen2-VL-2B-Instruct \cite{wang2024qwen2vlenhancingvisionlanguagemodels} as our image description backbone, and use DINOv3 (ConvNeXt-Tiny) \cite{simeoni2025dinov3} as our image encoder. See supplementary for more details.

\myheading{Inference. } On seen split, we estimate a subject's representation by sampling up to $K=100$ (see supplementary for choice of $K$) image-description-trajectory triplets from the training set of the same subject. Unseen split follows the same way, but using the support set. Then we compute a DEPER embedding for each triplet of a subject and average them to obtain a single subject embedding for this subject. 
This design removes the need to use trajectories for test images, which simplifies deployment and helps in practice. 

\begin{table*}[t]
\caption{Quantitative results on four datasets of \textbf{seen} subjects split. The first two baselines are population-level image description models, and the others are personalization models. B1=BLEU-1, B4=BLEU-4, M=METEOR, R=ROUGE\textsubscript{L}, C=CIDEr, P=Polos. Best per dataset in \textbf{bold}. \label{tab:seen-main-result}}
\centering
\scriptsize
\setlength{\tabcolsep}{3pt}
\renewcommand{\arraystretch}{1.12}

\resizebox{0.90\textwidth}{!}{
\begin{tabular}{c|cccccccc|cccccccc}
\toprule
\multirow{2}{*}{\textbf{Method}} &
\multicolumn{8}{c|}{\textbf{COCO-LN}~\cite{pont2020connecting}} &
\multicolumn{8}{c}{\textbf{Flickr30k-LN}~\cite{pont2020connecting}} \\
& \textbf{B1} & \textbf{B4} & \textbf{M} & \textbf{R} & \textbf{C} & \textbf{P} & \textbf{OSS} & \textbf{CLS}
& \textbf{B1} & \textbf{B4} & \textbf{M} & \textbf{R} & \textbf{C} &\textbf{P}
& \textbf{OSS} &\textbf{CLS}\\
\midrule
MITR~\cite{meng2021connecting} & 0.415 & 0.142 & 0.182 & 0.336 & 0.139 & 0.512 & 0.218 & -- & 0.296 & 0.076 & 0.160 & 0.281 & 0.077 & 0.319 & 0.200 & --\\
Qwen Zero-shot~\cite{wang2024qwen2vlenhancingvisionlanguagemodels} & 0.161 & 0.023 & 0.165 & 0.181 & 0.007 & 0.584 & 0.115 & -- & 0.169 & 0.024 & 0.166 & 0.187 & 0.004 & 0.612 & 0.133 & -- \\
\cmidrule(lr){1-17}
MITR-FT~\cite{meng2021connecting} & 0.437 & 0.176 & 0.209 & 0.341 & 0.142 & 0.589 & 0.246 & 0.415 & 0.302 & 0.101 & 0.169 & 0.298 & 0.094 & 0.320 & 0.224 & 0.427\\
CSMN \cite{chunseong2017attend} & 0.295 & 0.086 & 0.156 & 0.299 & 0.086 & 0.177 & 0.133 & 0.443 & 0.071 & 0.010 & 0.042 & 0.083 & 0.003 & 0.142 & 0.070 & 0.459 \\
Qwen+PT & 0.304 & 0.145 & 0.205 & 0.432 & 0.587 & 0.612 & 0.340 & 0.623 & 0.271 & 0.135 & 0.207 & 0.426 & 0.498 & 0.654 & 0.320 & 0.563\\
\textbf{Qwen+DEPER (Ours)} & \textbf{0.510} & \textbf{0.264} & \textbf{0.240} & \textbf{0.482} & \textbf{0.726} & \textbf{0.638} &\textbf{0.392} & \textbf{0.686} & \textbf{0.542} & \textbf{0.312} & \textbf{0.272} & \textbf{0.518} & \textbf{0.789} & \textbf{0.671} & \textbf{0.408} & \textbf{0.796} \\
\midrule
\end{tabular}
}

\vspace{3pt}

\resizebox{0.90\textwidth}{!}{
\begin{tabular}{c|cccccccc|cccccccc}
\midrule
\multirow{2}{*}{\textbf{Method}} &
\multicolumn{8}{c|}{\textbf{Kollenda \etal}~\cite{kollenda2025individual}} &
\multicolumn{8}{c}{\textbf{He \etal}~\cite{he2019human}} \\
& \textbf{B1} & \textbf{B4} & \textbf{M} & \textbf{R} & \textbf{C} & \textbf{P} & \textbf{OSS} & \textbf{CLS}
& \textbf{B1} & \textbf{B4} & \textbf{M} & \textbf{R} & \textbf{C} &\textbf{P}
& \textbf{OSS} &\textbf{CLS}\\
\midrule
Qwen Zero-shot~\cite{wang2024qwen2vlenhancingvisionlanguagemodels} & 0.267 & 0.047 & 0.135 & 0.291 & 0.363 & 0.559 & 0.317 &-- & 0.430 & 0.132 & 0.209 & 0.422 & 1.144 & 0.578 & 0.377 & -- \\
\cmidrule(lr){1-17}
CSMN \cite{chunseong2017attend} & 0.019 & 0.001 & 0.012 & 0.034 & 0.002 & 0.100 & 0.012 & 0.021 & 0.025 & 0.008 & 0.032 & 0.003 & 0.004 & 0.091 & 0.015 & 0.157 \\
Qwen+PT & 0.344 & 0.067 & 0.176 & 0.325 & 0.504 & 0.527 & 0.328 & 0.056 & 0.475 & 0.174 & 0.211 & 0.446 & 1.515 & 0.589 & 0.448 & 0.262\\
\textbf{Qwen+DEPER (Ours)} & \textbf{0.442} & \textbf{0.135} & \textbf{0.201} & \textbf{0.382} & \textbf{0.871} & \textbf{0.594} & \textbf{0.351} &\textbf{0.083} &\textbf{0.506} & \textbf{0.207} & \textbf{0.234} & \textbf{0.486} & \textbf{1.822} & \textbf{0.603} & \textbf{0.470} &\textbf{0.307}\\
\midrule
\end{tabular}
}
\vskip -0.1in
\end{table*}

\subsection{Evaluation Metrics}
We adopt two different kinds of metrics to conduct a comprehensive evaluation.

\myheading{Widely adopted captioning metrics.}  We use BLEU-1, BLEU-4 \cite{papineni2002bleu}, METEOR \cite{banerjee2005meteor}, ROUGE\textsubscript{L} \cite{lin2004rouge}, CIDEr \cite{vedantam2015cider}. We also use Polos \cite{wada2024}, a recently proposed metric using a pretrained vision-language model to evaluate image captioning performance.

\myheading{Personalization-centric evaluation}. Following \cite{yang2024unifying,chakraborty2025measuring}, we propose the Object Sequence Score (\textbf{OSS}): an object-level metric that compares \emph{which} objects are mentioned and their \emph{order}, capturing personalized narrative alignment. We extract ordered nouns from the prediction and reference, then align them via Needleman--Wunsch \cite{needleman1970} with weighted matches of exact, stem and synonym.
We also implement top-1 classification accuracy (\textbf{CLS}), to assess whether a subject’s generated description is distinguishable from other subjects for the same image. We rank each generated description against all other subjects’ generated descriptions for images with $\geq$ 3 subject descriptions (Flickr30k-LN $\geq$ 2  because it has few examples with more than 2 ground-truth descriptions). A hit is when the same-subject description ranks first under a metric; we report the mean hit rate over the BLEU-4, METEOR, ROUGE-L, and CIDEr metrics.

\subsection{Main Results}

\myheading{Human Consistency. }To quantify how differently humans describe the same image, we compute Human Consistency (HC) using \textit{m-BLEU-4} \cite{mahajan2020diverse}, which measures the average similarity among descriptions of the same image, with units comparison to BLEU-4. As shown in \cref{tab:data-stats}, human descriptions vary substantially. Identical captions yield HC = 1, while lower scores indicate greater diversity.

\myheading{Performance on seen subjects.} As shown in \cref{tab:seen-main-result}, our method achieves the best performance on all metrics, showing improvements (averaged over datasets) of 62\% on BLEU-4, 28\% on CIDEr, 13\% on OSS, and 29\% on CLS. The improvement from Qwen+PT to Qwen+DEPER shows that, beyond simple prompt tuning, DEPER learns subject embeddings that more faithfully capture each author’s viewing patterns and linguistic style, producing higher-quality personalized descriptions.

Our method also yields strong gains on personalization-focused metrics. The 13.0\% OSS gain reflects better modeling of the three key components of human attention: which objects are mentioned, in what order, and to what amount of detail. The 15.4\% CLS gain shows that our generated descriptions effectively capture subjects' distinctive features. The 30-way classification on Kollenda \etal is hard, yet our method achieves a $\sim$150\% improvement over chance. CLS is not computed for population-level models, which generate identical descriptions for an image regardless of the subject.

\begin{table}[t]
\caption{Personalized description generation performance on \textbf{unseen} subjects. C\cite{pont2020connecting}=COCO-LN, F\cite{pont2020connecting}=Flickr30k-LN\cite{pont2020connecting}, K\cite{kollenda2025individual}=Kollenda \etal\cite{kollenda2025individual}. Best per dataset in \textbf{bold}. \label{tab:unseen-main-results}}
\centering
\small
\setlength{\tabcolsep}{1.2pt}
\renewcommand{\arraystretch}{1.2}
\resizebox{1.0\columnwidth}{!}{
\begin{tabular}{c|c|ccccccc}
\toprule
\textbf{Dataset} & \textbf{Method} & \textbf{B4} & \textbf{M} & \textbf{R}  & \textbf{C} & \textbf{P} &\textbf{OSS} & \textbf{CLS}\\
\midrule
\multirow{4}{*}{\textbf{C}\cite{pont2020connecting}} 
  & MITR-FT\cite{park2018towards} & 0.139 & 0.179 & 0.334 & 0.135 & 0.491 & 0.201 & 0.326\\
  & CSMN\cite{chunseong2017attend} & 0.042 & 0.105 & 0.133 & 0.026 & 0.149 & 0.198 & 0.312 \\
  & Qwen few-shot\cite{wang2024qwen2vlenhancingvisionlanguagemodels} & 0.071 & 0.106 & 0.203 & 0.077 & 0.397 & 0.142 & 0.406 \\
  & Qwen+PT & 0.058 & 0.141 & 0.265 & 0.317 & 0.434 & 0.138 & 0.317\\
  & \textbf{Ours} & \textbf{0.164} & \textbf{0.184} & \textbf{0.389} & \textbf{0.453} & \textbf{0.597} &\textbf{0.330} & \textbf{0.445}\\
  \midrule
  \multirow{4}{*}{\textbf{F}\cite{pont2020connecting}}
  & MITR-FT\cite{park2018towards} & 0.074 & 0.126 & 0.261 & 0.104 & 0.209 & 0.187 & 0.415 \\
  & CSMN\cite{chunseong2017attend} & 0.007 & 0.021 & 0.053 & 0.001 & 0.102 & 0.029 & 0.427 \\
  & Qwen few-shot\cite{wang2024qwen2vlenhancingvisionlanguagemodels} & 0.122 & 0.148 & 0.241 & 0.068 & 0.386 & 0.150 & 0.416 \\
  & Qwen+PT & 0.074 & 0.167 & 0.337 & 0.338 & 0.587 & 0.278 & 0.479 \\
  & \textbf{Ours} & \textbf{0.202} & \textbf{0.232} & \textbf{0.410} & \textbf{0.382} & \textbf{0.610} & \textbf{0.329} &\textbf{0.625} \\
\midrule
\multirow{4}{*}{\textbf{K}\cite{kollenda2025individual}}
  & CSMN\cite{chunseong2017attend} & 0.003 & 0.007 & 0.015 & 0.000 & 0.092 & 0.004 & 0.085\\
  & Qwen few-shot\cite{wang2024qwen2vlenhancingvisionlanguagemodels} & 0.063 & 0.157 & 0.288 & 0.538 & 0.535 & 0.272 & 0.151 \\
  & Qwen+PT & 0.019 & 0.148 & 0.237 & 0.209 & 0.507 & 0.220 & 0.111 \\
  & \textbf{Ours} & \textbf{0.143} & \textbf{0.207} & \textbf{0.398} & \textbf{1.053} & \textbf{0.583} & \textbf{0.380} & \textbf{0.157}\\
\bottomrule
\end{tabular}}
\vskip -0.1in
\end{table}

\begin{figure*}[h]
  \centering
  \includegraphics[width=1.0\linewidth]{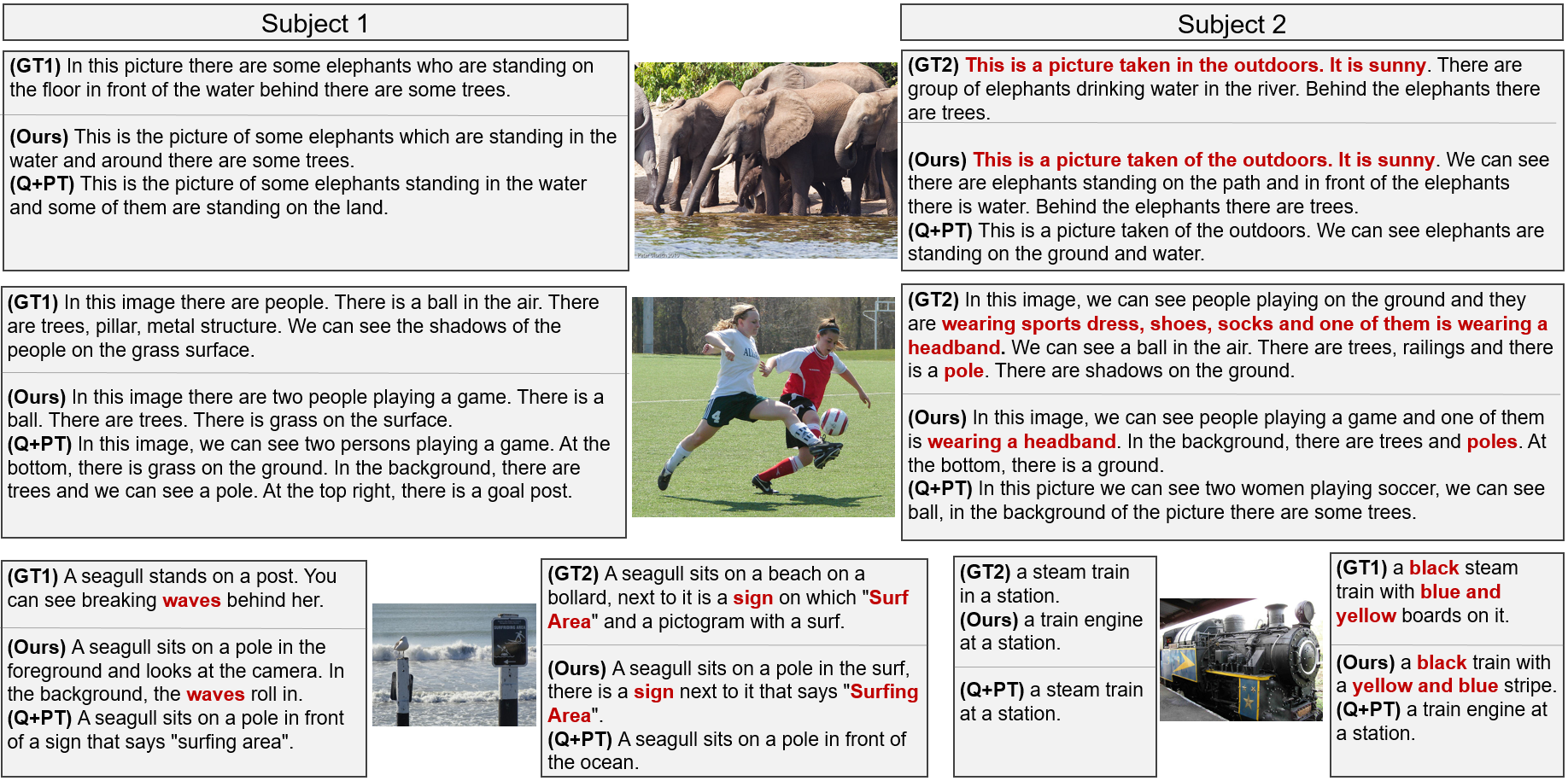}
  \caption{\textbf{Qualitative Results} show one example per dataset, each with two subject-specific descriptions (subjects 1 and 2). From top to bottom and left to right: COCO-LN~\cite{pont2020connecting}, Flickr30k-LN~\cite{pont2020connecting}, Kollenda \etal~\cite{kollenda2025individual}, and He \etal~\cite{he2019human}. Subject-distinct content is highlighted in \textcolor{red}{red}. Qwen+PT is denoted as Q+PT.}
  \label{fig:main-result}
  \vspace{-1em}
\end{figure*} 
\myheading{Qualitative Results on seen subjects. }In \cref{fig:main-result}, we show one example per dataset, each with two subjects and their brief or detailed descriptions. These cases illustrate components of subject-specific variation that our method captures: (1) object-centric vs. scene-centric openings (Example 1); (2) description granularity, where one subject offers fine-grained details while the other is terse (Examples 2 and 4); and (3) object-of-interest selection, e.g., background-oriented vs. sign-focused attention (Example 3).

\myheading{Performance on unseen subjects.} In \cref{tab:unseen-main-results}, our method shows strong few-shot performance across all datasets. On Kollenda \etal, the unseen split exhibits only a small drop from the seen split. Few-shot personalization is harder on COCO-LN and Flickr30k-LN because their support images differ from the seen-subject training data; yet, our performance remains stable across all five metrics. The higher CLS scores indicate that DEPER can infer distinct subject embeddings for previously unseen individuals, demonstrating the effective transfer of subject-specific cues. Notably, our method extracts subject embeddings without per-subject fine-tuning, enabling real-time adaptation and avoiding separate models for each subject. We omit unseen evaluation on He \etal due to its limited subject count (N=5).


\begin{figure}[t]
  \centering
  \includegraphics[width=1\linewidth]{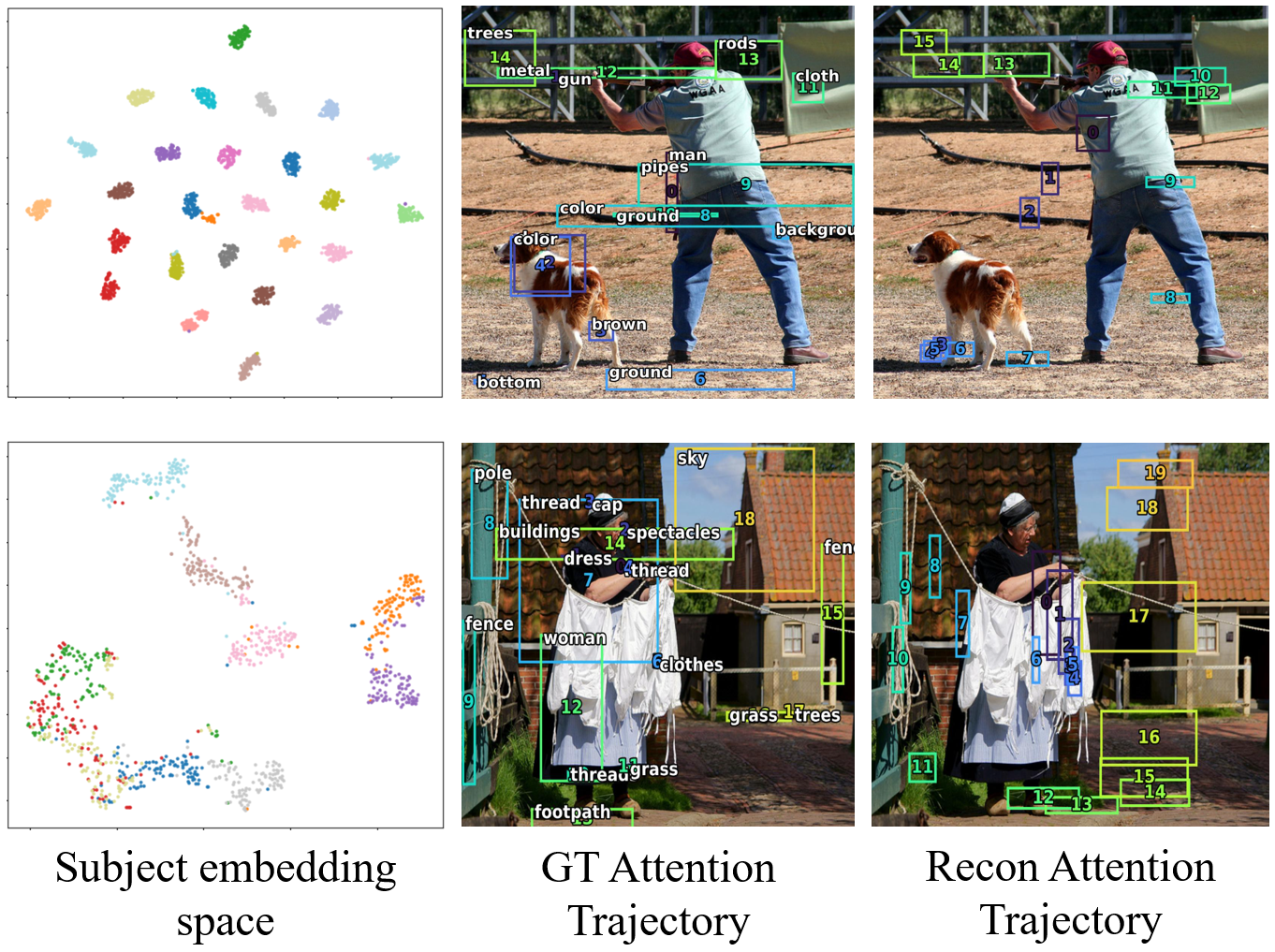}
  \caption{\textbf{Qualitative results of DEPER's outputs.} We show DEPER’s outputs on seen and unseen splits (first and second rows) of Flickr30k-LN. The first column visualizes DEPER's subject embeddings, where colors denote subjects and each point represents an image–description–trajectory triplet. The second column shows ground-truth attention trajectories with their corresponding nouns and orders; the third column shows reconstructed trajectories from the test set after Stage-2 training.}
  \label{fig:cls-and-recon}
  \vspace{-1em}
\end{figure}
\myheading{Qualitative results on DEPER's outputs. } \cref{fig:cls-and-recon} visualizes DEPER’s learned subject embeddings and reconstructed attention trajectories for both seen and unseen subjects. The embedding space forms clear, subject-specific clusters, while the ground-truth and reconstructed trajectories illustrate that DEPER captures visual dynamics well. The coherent clustering of unseen subjects and the close match between reconstructed and true trajectories highlight strong generalization to new individuals. We present further results in supplementary.

\subsection{Ablations}
\begin{table}[h!]
    \caption{\textbf{Ablation on DEPER's components.}
    \textit{Text Input} and \textit{Traj Input} denote the input modalities provided to DEPER, \textit{Traj Dyn} denotes the duration and index of each bounding box. \textit{Traj Recon} refers to the presence of the trajectory–reconstruction decoder module. \checkmark indicates the inclusion of the corresponding module. The results show the importance of attention trajectories in image description generation at different stages. Results are reported on the Flickr30k-LN.}
    \centering
    \renewcommand{\arraystretch}{1.05} 
    \setlength{\tabcolsep}{1pt} 
    \resizebox{1.0\columnwidth}{!}{
    \begin{tabular}{
        >{\centering\arraybackslash}p{1.0cm}
        >{\centering\arraybackslash}p{1.0cm}
        >{\centering\arraybackslash}p{1.0cm}
        >{\centering\arraybackslash}p{1.0cm}|
        >{\centering\arraybackslash}p{1.2cm}
        >{\centering\arraybackslash}p{1.2cm}
        >{\centering\arraybackslash}p{1.2cm}
        >{\centering\arraybackslash}p{1.2cm}
        >{\centering\arraybackslash}p{1.2cm}
        >{\centering\arraybackslash}p{1.2cm}
        >{\centering\arraybackslash}p{1.2cm}
    }
        \toprule
        \textbf{Text Input} & \textbf{Traj Input} &\textbf{Traj Dyn}& \textbf{Traj Recon} & \textbf{B4} & \textbf{M} & \textbf{R} & \textbf{C} & \textbf{OSS} &\textbf{CLS}\\
        \midrule
        \checkmark & -- & -- & -- & 0.222 & 0.247 & 0.500 & 0.770 & 0.379 & 0.649\\
        \checkmark & \checkmark & -- & \checkmark & 0.276 & 0.266 & 0.514 & 0.748 & 0.378 & 0.731 \\
        \checkmark & \checkmark & \checkmark & -- & 0.230 & 0.256 & 0.499 & 0.774 & 0.381 & 0.724\\
        \checkmark & \checkmark & \checkmark & \checkmark & \textbf{0.312} & \textbf{0.272} & \textbf{0.518} & \textbf{0.789} & \textbf{0.408} &\textbf{0.796}\\
        \bottomrule
    \end{tabular}
    }
    \vskip -0.05in
    \label{tab:ablation-module}
\end{table}

\myheading{Ablation on attention trajectory.}
To evaluate the role of human attention in shaping subject embeddings, we ablate three components in \cref{tab:ablation-module}: adding attention-trajectory features, adding visual dynamics (fixation duration and order), and adding the trajectory reconstruction objective. Removing attention trajectories entirely causes a clear drop in performance, showing that human attention is critical to modeling subject-specific perception.

The results also show that multiple elements of attention are needed to achieve full performance. An ablation omitting trajectory dynamics (duration and order) drops performance to a halfway point between the no-attention ablation and the full model, highlighting the key role of attention dynamics in human scene description. An ablation removing the trajectory-reconstruction objective fares even worse, showing the importance of preserving attentional information in $\mathbf{Z}_{dual}$ and $\mathbf{z}_s$ rather than encoding linguistic style alone.

\begin{table}[h!]
\caption{\textbf{Ablation on the effect of different modules}. We report performance under key removals on Flickr30k-LN: w/o Traj Latent (directly use subject embedding to reconstruct trajectory), w/o Dual-Context (dual-context encoder removed), w/o Contrast (contrastive loss disabled in both stages), w/o FiLM (no subject-embedding FiLM modulation in trajectory decoder). \label{tab:ablation-loss}} 
    \centering
    \renewcommand{\arraystretch}{1} 
    \setlength{\tabcolsep}{4pt} 
    
    \resizebox{0.85\columnwidth}{!}{
    \begin{tabular}{lcccccc}
        \toprule
         Modules& B4 & M & R & C & OSS & CLS\\
        \midrule
        w/o Dual-Context & 0.229 & 0.252 & 0.495 & 0.729 & 0.380 & 0.731\\
        w/o Traj Latent & 0.272 & 0.261 & 0.505 & 0.745 &0.391 & 0.750 \\
        w/o Contrast & 0.228 & 0.259 & 0.498 & 0.743 & 0.386 & 0.722 \\
        w/o FiLM & 0.270 & 0.262 & 0.506 & 0.723 & 0.394 & 0.768 \\
        Ours & 0.312 & 0.272 & 0.518 & 0.789 & 0.408 & 0.796 \\
        \bottomrule
    \end{tabular}
    }
    \vskip -0.1in        
\end{table}
\myheading{Ablation on modules. }We further ablate different modules in \cref{tab:ablation-loss}. Removing any module leads to a significant drop, indicating that each supplies a complementary signal: the Dual-Context Encoder learns cross-modality correspondences; the trajectory latent $\mathbf{z}_{traj}$ alleviates the need for the subject embedding to encode instance-specific information needed for reconstruction; the contrastive loss encourages DEPER to distinguish subject embeddings; and FiLM modulates the trajectory reconstruction with a subject-aware signal.

\subsection{Analysis}

\begin{table}[h!]
    \caption{\textbf{Performance with varying training sizes.} \# Samples denotes the number of training examples per subject. The results show DEPER's data efficiency. Results are reported on the Flickr30k-LN dataset. \label{tab:num-train-per-subj} }   
    \centering
    \renewcommand{\arraystretch}{1} 
    \setlength{\tabcolsep}{4pt} 
    
    \resizebox{0.80\columnwidth}{!}{
    \begin{tabular}{l|cccccc}
        \toprule
        {\# samples} & \textbf{B4} & \textbf{M} & \textbf{R} & \textbf{C} & \textbf{OSS} & \textbf{CLS}\\
        \midrule
        100 & 0.171 & 0.200 & 0.402 & 0.400 & 0.381 & 0.730 \\
        200 & 0.268 & 0.255 & 0.497 & 0.664 & 0.383 & 0.750 \\
        500 & 0.292 & 0.271 & 0.513 & 0.754 & 0.389 & 0.758\\
        801(Full) & 0.312 & 0.272 & 0.518 & 0.789 & 0.408 & 0.796\\
        \bottomrule
    \end{tabular}
    }
    \vskip -0.1in
\end{table}
\textbf{Data efficiency analysis.}
To assess DEPER’s dependence on training set size, we vary the number of training samples per subject and test performance on the seen split. As shown in \cref{tab:num-train-per-subj}, performance drops only slightly when using 62\% of the full training set, showing strong data efficiency. Even with as few as 100 samples per subject (2,700 total), DEPER maintains performance comparable to baselines trained on the complete dataset. These results highlight DEPER’s ability to learn robust subject embeddings from limited supervision, which substantially reduces data requirements for VLM fine-tuning and enables applications in data-scarce domains such as healthcare and assistive vision.

\section{Conclusion and Discussion}
We study how personalized human attention shapes personalized image descriptions, noting that people describe the image differently not only in what they choose to say and what words they use, but also in the order and detail in which they continuously explore a scene. However, modeling this implicit and consistent viewing behavior is challenging, especially when a system must adapt to new subjects with only a few examples. To address this, we develop a unified pipeline that learns a subject-specific representation capturing both a person’s visual exploration behavior and linguistic style, and uses this representation to guide a frozen VLM to generate personalized descriptions. 
DEPER consistently outperforms prior methods on seen subjects and achieves strong few-shot generalization to unseen individuals. In ablations, we demonstrate that attention signals are critical to this performance. Altogether, our approach sets a new state-of-the-art in personalized image description generation task and highlights the value of richer, behavior-aware subject representations for future research.\\
Future work may extend DEPER to learn personality representations from tasks beyond image description. Attention–language alignment also affects performance in visual question answering, where individuals rely on different visual cues, and could further support robotics applications by helping robots infer the implicit factors that guide human behavior.
{
    \small
    \bibliographystyle{ieeenat_fullname}
    \bibliography{main}
}

\end{document}